\documentclass[10pt]{article}

\usepackage[utf8]{inputenc}
\usepackage[T1]{fontenc}
\usepackage[top=2cm, bottom=2.2cm, left=2.5cm, right=2.5cm]{geometry}
\usepackage{amsmath}
\usepackage{amssymb}

\usepackage{graphicx}
\usepackage{xcolor}

\definecolor{primaryblue}{RGB}{20, 80, 160}
\definecolor{lightblue}{RGB}{220, 235, 250}
\definecolor{midblue}{RGB}{160, 200, 240}
\definecolor{accentorange}{RGB}{220, 100, 30}
\definecolor{lightorange}{RGB}{250, 225, 200}
\definecolor{lightgray}{RGB}{245, 245, 247}
\definecolor{medgray}{RGB}{210, 210, 215}
\definecolor{darkgray}{RGB}{80, 80, 80}
\definecolor{successgreen}{RGB}{40, 150, 80}
\definecolor{nodepurple}{RGB}{100, 80, 200}
\definecolor{nodecoral}{RGB}{200, 80, 60}
\definecolor{nodeteal}{RGB}{20, 140, 110}
\definecolor{nodeblue}{RGB}{30, 100, 200}

\usepackage{booktabs}
\usepackage{tabularx}
\usepackage{array}
\usepackage{enumitem}
\usepackage{parskip}

\newcolumntype{Y}{>{\centering\arraybackslash}X}

\usepackage{tikz}
\usetikzlibrary{arrows.meta, positioning, calc}

\usepackage{pgfplots}
\pgfplotsset{compat=1.17}
\usepgfplotslibrary{groupplots}

\usepackage{tcolorbox}
\tcbuselibrary{skins, breakable}

\usepackage{titlesec}
\usepackage{fancyhdr}

\titleformat{\section}{\large\bfseries}{\color{primaryblue}\thesection}{1em}{\color{black}}
\titleformat{\subsection}{\normalsize\bfseries}{\color{primaryblue}\thesubsection}{1em}{\color{black}}
\titlespacing*{\section}{0pt}{10pt}{8pt}
\titlespacing*{\subsection}{0pt}{6pt}{6pt}

\usepackage{float}
\usepackage[section]{placeins}
\usepackage{etoolbox}
\pretocmd{\subsection}{\FloatBarrier}{}%
  {\PackageWarning{gra}{Could not patch \string\subsection\space with \string\FloatBarrier}}

\usepackage[numbers,square,sort&compress]{natbib}

\usepackage{tocloft}

\usepackage{hyperref}
\hypersetup{colorlinks=true, linkcolor=primaryblue, urlcolor=primaryblue,
            citecolor=primaryblue, linktoc=all}

\usepackage{caption}
\newtcolorbox{examplebox}[2][]{%
  enhanced, breakable,
  colback=lightblue!22, colframe=primaryblue!28,
  boxrule=0.55pt, arc=5pt,
  left=7pt, right=7pt, top=7pt, bottom=7pt,
  title={#2},
  fonttitle=\bfseries\color{primaryblue},
  coltitle=primaryblue, #1
}

\newcommand{\ci}[1]{{\scriptsize\color{darkgray}$\pm$#1}}
\newcommand{\rd}[1]{\makebox[2.7em][r]{#1}}

\begin{document}

\thispagestyle{empty}

\begin{tcolorbox}[colback=lightgray, colframe=lightgray, arc=10pt, boxrule=0pt,
  left=20pt, right=20pt, top=18pt, bottom=18pt]

\includegraphics[height=26pt]{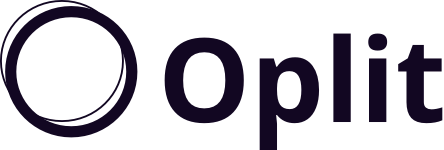}\\[8pt]

{\fontsize{23}{29}\selectfont\bfseries\color{primaryblue} Schema-Agnostic Graph Reasoning
Agent for Hybrid Knowledge Graphs}\\[3pt]

\vspace{8pt}

{\normalsize\color{darkgray}
\textbf{Marius Dragic}\textsuperscript{*}\quad \textbf{Alexandre Rio}\textsuperscript{*}\quad \textbf{Ruben Ifrah}\textsuperscript{*}}\\[2pt]
{\small\color{darkgray}
R\&D Department, Oplit}\\[2pt]
{\small\color{primaryblue}\texttt{\{marius, alexandre.rio, ruben\}@oplit.fr}}

\vspace{8pt}
\textcolor{medgray}{\hrule}
\vspace{10pt}

{\small\color{darkgray}
Tool-calling LLM agents navigate unfamiliar codebases with a handful of generic primitives for listing, reading and searching files (\texttt{ls}, \texttt{cat}, \texttt{grep}). A knowledge graph admits the same interface: listing neighbours, reading node content and searching descriptions are the same operations on a different substrate. Building on this correspondence, we present \textbf{GRA}, a Graph Reasoning Agent that explores \emph{hybrid} knowledge graphs, whose nodes are either textual concepts or relational tables, with seven generic tools, discovering everything domain-specific at run time. On \textbf{UFK-M} (Unified Factory Knowledge Model), an industrial benchmark of 258 analytical questions whose gold answers are produced by executing validated SQL programs, GRA beats a full-context agent by \textbf{5.1 pp} (88.4\% vs.\ 83.3\%), while reading under a third of its input tokens. A graph-free control shows the gain comes chiefly from selective agentic access rather than graph topology, and that the effect depends on a model able to drive tools reliably. Seeing less, the agent answers better: selective navigation
over a structured substrate beats exhaustive context.
}
\vspace{6pt}
{\scriptsize\color{darkgray}
\textbf{Date:} July 2026 \hfill \textbf{*Equal contribution}}

\end{tcolorbox}

\begin{tcolorbox}[colback=white, colframe=medgray,
  arc=8pt, boxrule=0.6pt,
  left=6pt, right=6pt, top=8pt, bottom=6pt,
  width=\linewidth]
  \begin{center}
  \begin{tikzpicture}[
    sem/.style={rounded corners=3pt, draw=#1, fill=#1!12, line width=0.9pt,
                align=center, font=\tiny, inner sep=3.5pt},
    agent/.style={rounded corners=6pt, draw=#1, line width=1pt, fill=#1!5,
                  minimum width=4.7cm, minimum height=1.6cm, align=center,
                  inner sep=4pt, font=\tiny},
    panel/.style={draw=medgray, line width=0.6pt, fill=white, rounded corners=6pt},
    tblh/.style={draw=darkgray!70, fill=midblue, minimum width=2.9cm,
                 minimum height=0.4cm, inner sep=1.5pt,
                 font=\tiny\ttfamily\bfseries, text=darkgray},
    tblb/.style={draw=darkgray!70, fill=white, minimum width=2.9cm,
                 inner sep=2.6pt, font=\tiny\ttfamily, align=left},
    edgeL/.style={font=\tiny, text=darkgray, fill=white, inner sep=1.5pt},
    arr/.style={-{Stealth[length=5pt]}, line width=0.9pt, darkgray!80},
    feed/.style={-{Stealth[length=5pt]}, line width=1pt, draw=#1}
  ]

  \draw[draw=darkgray!55, line width=0.9pt, fill=lightgray, rounded corners=10pt]
      (-7.55,-0.45) rectangle (7.55,3.70);
  \node[font=\small\bfseries, text=darkgray, anchor=west] at (-7.30,3.30)
      {UFK-M --- Unified Factory Knowledge Model};

  \draw[panel] (-7.30,0.05) rectangle (-1.20,3.00);
  \node[font=\scriptsize\bfseries, text=darkgray, anchor=west] at (-7.10,2.74)
      {Semantic layer};

  \node[sem=nodeteal]   (rule) at (-6.00,2.15) {\textbf{Operational}\\ \textbf{Rule} R8};
  \node[sem=nodeblue]   (cust) at (-2.75,2.15) {\textbf{Concept}\\ Customer};
  \node[sem=nodepurple] (kpi)  at (-6.00,0.90) {\textbf{KPI}\\ Lateness};

  \draw[panel] (-0.60,0.05) rectangle (7.30,3.00);
  \node[font=\scriptsize\bfseries, text=darkgray, anchor=west] at (-0.40,2.74)
      {Data layer};

  \node[tblh] (t1h) at (1.80,2.15) {tbl\_customers};
  \node[tblb, anchor=north] (t1b) at (t1h.south)
      {customer\_id (pk)\\ segment~$\cdot$~market};
  \node[tblh] (t2h) at (5.45,1.10) {tbl\_orders};
  \node[tblb, anchor=north] (t2b) at (t2h.south)
      {order\_id (pk)~$\cdot$~cust\_id (fk)\\ promised~$\cdot$~delivered};

  \draw[arr] (rule.east) -- (cust.west) node[edgeL, midway]{CONSTRAINS};
  \draw[arr, dashed] (cust.east) -- (t1h.west) node[edgeL, midway]{REPRESENTS};
  \draw[arr, rounded corners=5pt] (t1b.east) -- (t1b.east -| t2h.north)
      node[edgeL, midway]{JOINS} -- (t2h.north);
  \draw[arr, rounded corners=5pt] (kpi.south) -- (-6.00,0.35) -- (3.55,0.35)
      -- (3.55,1.10) -- (t2h.west);
  \node[edgeL] at (-3.75,0.35) {MEASURED\_ON};

  \node[agent=primaryblue]  (gra) at (-5.1,-2.48) {%
    {\footnotesize\bfseries\color{primaryblue}GRA --- graph agent}\\[3pt]
    \texttt{ls~$\cdot$~cat~$\cdot$~grep~$\cdot$~sems}\\
    \texttt{query~$\cdot$~think~$\cdot$~answer}\\[3pt]
    {\color{darkgray}schema discovered at run time}};
  \node[agent=successgreen] (rsa) at (0,-2.48) {%
    {\footnotesize\bfseries\color{successgreen}RSA --- flat agent}\\[3pt]
    \texttt{search\_text~$\cdot$~describe\_table}\\
    \texttt{query~$\cdot$~think~$\cdot$~answer}\\[3pt]
    {\color{darkgray}same loop, graph removed}};
  \node[agent=accentorange] (sqa) at (5.1,-2.48) {%
    {\footnotesize\bfseries\color{accentorange}SQA --- full context}\\[3pt]
    founding text + full schema\\
    ($\sim$17\,k tokens up front)\\[3pt]
    {\color{darkgray}$\le$6 turns, no navigation}};

  \draw[feed=primaryblue]  (-5.1,-0.45) -- (gra.north)
      node[edgeL, midway, text=primaryblue]{reads graph slices};
  \draw[feed=successgreen] (0,-0.45) -- (rsa.north)
      node[edgeL, midway, text=successgreen]{retrieves chunks + schemas};
  \draw[feed=accentorange] (5.1,-0.45) -- (sqa.north)
      node[edgeL, midway, text=accentorange]{serialized into the prompt};

  \end{tikzpicture}
  \end{center}
  \vspace{2pt}\textcolor{medgray}{\hrule}\vspace{5pt}
    {\scriptsize\color{darkgray}
      One substrate, three tested baselines. A semantic layer
      of concepts, KPIs and rules is bridged to relational table nodes; GRA
      navigates it with generic tools, RSA is the same agent with the graph
      removed, and SQA receives everything serialized in its prompt.}
\end{tcolorbox}


\newpage

\begingroup
  \hypersetup{linkcolor=black}   
  \setlength{\parskip}{0pt}      
  \tableofcontents
\endgroup

\newpage

\section{From code agents to graph agents}

Tool-calling agents work effectively inside repositories they have never
seen, using a small set of generic primitives: list a directory, read a
file, search for a string. None of these tools knows anything about the
project; the agent reconstructs it by navigating. A labeled property graph
admits the same interface. Listing the neighbours of a node, reading a
node's content, and searching node descriptions are structurally equivalent
operations on a different substrate, so the navigation competence of code
agents should transfer to knowledge graphs at little cost.

The substrate contract we assume is minimal. Every node carries an
identifier, a natural-language description, labels, and optional
properties; relations are directed subject--predicate--object triples. Some
nodes are \emph{data tables} and are backed by a real DuckDB table that can
be queried with SQL. This is the \emph{hybrid} in hybrid knowledge graph:
textual and relational knowledge coexist in one graph, and answering a
question may require either or both. Two philosophies compete to feed
such a substrate. The first serializes everything (documentation
and schema) into the prompt and asks for an answer. The second gives the
model a bounded context and a toolset, and lets it fetch what it needs.
This paper measures that trade-off, with controls designed to attribute the
outcome to its cause.

\section{Related works}
GRA's interface descends from ReAct~\citep{yao2023react}, which introduced
the interleaving of reasoning and tool calls now common to LLM agents but
left open which tools a given substrate should expose. For source code,
SWE-agent~\citep{yang2024sweagent} answered that question: its
\emph{agent--computer interface} showed that a few file-system commands
suffice to work over an unfamiliar repository. A knowledge graph is much
like a codebase in this respect --- a large connected structure explored by
following links, not by reading it whole --- so the strategy transfers
directly; only the substrate differs.

A parallel line gives language models an interface to structured knowledge
directly, as surveys of graph retrieval-augmented generation
describe~\citep{peng2024graphrag}. These systems read over graphs, tables,
and databases~\citep{jiang2023structgpt}, follow relation
paths~\citep{sun2024tog, ma2025tog2, search2024sog}, or expose primitives
for node lookup and neighbour listing~\citep{jin2024graphcot,
jiang2024kgagent}. All share GRA's premise --- navigate the graph rather
than read a flat dump of it --- but assume that the graph's vocabulary is
known and that traversal alone answers the question. Neither assumption
holds on a hybrid substrate, where the deciding fact often sits in a table
reached through the graph, and where a question's words must first be
matched to a node; aligning the two is itself hard, as work on
natural-language-to-graph-query translation
shows~\citep{liang2024nl2gql}. GraphRAG~\citep{edge2024graphrag} instead
builds a graph and summaries offline and retrieves over them, which suits
broad questions but fixes the retrieval structure in advance, whereas GRA
gathers evidence per question over a graph that keeps changing.

Each of these provides part of what GRA needs, but none provides all of it
at once: operation without per-schema tuning, a substrate that unites a
semantic graph with relational tables, and the ability to compute a
quantity no node stores. Section~\ref{sec:overview} puts that combination
to work, where turning an operator's plain-language rule into a grounded
feasibility verdict depends on all three together.

\section{Three agents, one substrate}

\subsection{GRA: Graph Reasoning Agent}
GRA explores the graph with the seven unix-style tools of Table~\ref{tab:tools}. Nothing domain-specific is hard-coded in the tools or the system prompt: no concept list, no label strings, no table names. The agent orients with \texttt{ls}, reads nodes with \texttt{cat}, searches literally with \texttt{grep} and semantically with \texttt{sems}, and reads values with a read-only \texttt{query}.
\vspace{0.5cm}

\begin{table}[H]
\centering
\small
\renewcommand{\arraystretch}{1.3}
\begin{tabularx}{\linewidth}{l X}
\toprule
\textbf{Tool} & \textbf{Role} \\
\midrule
\texttt{ls}    & List nodes, tables, and edges to orient in the graph. \\
\texttt{cat}   & Read a single node in full; for tables, shows columns, keys, joins, and a 3-row sample. \\
\texttt{grep}  & Literal search over ids, text, properties, edges, and columns. \\
\texttt{sems}  & Semantic search (dense + BM25, top 10 results). \\
\texttt{query} & Read-only SQL (\texttt{SELECT}/\texttt{WITH}), capped at 50 rows. \\
\texttt{think} & Scratchpad for planning and checking, no side effects. \\
\texttt{answer} & Submit the final answer, with optional citations and confidence. \\
\bottomrule
\end{tabularx}
\caption{The GRA toolkit: seven generic primitives for navigating and querying a hybrid knowledge graph.}
\label{tab:tools}
\end{table}

\begin{examplebox}{Example --- GRA on UFK-M}
\textbf{Question.} \emph{``How many orders were late for each customer
segment last quarter?''} The agent starts cold: no schema, no table names,
no vocabulary.\\[3pt]
\textbf{Trace.} \texttt{ls(labels)} $\rightarrow$ \texttt{ls(tables)}
$\rightarrow$ \texttt{grep("orders")} $\rightarrow$
\texttt{grep("customer segment")} $\rightarrow$
\texttt{cat(tbl\_customer\_orders)} $\rightarrow$ \texttt{cat(Customer)}
$\rightarrow$ \texttt{cat(tbl\_customers)} $\rightarrow$
\texttt{query(\dots)} $\rightarrow$ \texttt{think} $\rightarrow$
\texttt{answer}.\\[3pt]
\textbf{Outcome.} The KG
and the definition of ``late'' are discovered on the way, and only a few
thousand unique tokens are ever read.
\end{examplebox}

\vspace{0.5cm}

\subsection{RSA: Retrieval SQL Agent}
RSA is GRA with the graph removed: it retrieves over chunks of the flat textual documentation and over table schemas (\texttt{search\_text}, \texttt{list\_tables}, \texttt{describe\_table}) and keeps the identical \texttt{query}, \texttt{think} and \texttt{answer} tools. Fairness is enforced by construction: both agents run the very same execution loop, share their strategy prompt blocks verbatim, and differ only in substrate and toolset.

\begin{examplebox}{Example --- RSA on the same question}
\textbf{Trace.} \texttt{search\_text("customer segment")} $\rightarrow$
\texttt{list\_tables()} $\rightarrow$
\texttt{describe\_table(tbl\_customer\_orders)} $\rightarrow$
\texttt{describe\_table(tbl\_customers)} $\rightarrow$
\texttt{query(\dots)} $\rightarrow$ \texttt{think} $\rightarrow$
\texttt{answer}.\\[3pt]
\textbf{Outcome.} The same loop reaches the same answer, but the link
between the segment concept and the table that stores it must be inferred
from retrieved text chunks rather than read off an edge, and candidate
tables are scanned by name instead of followed.
\end{examplebox}

\subsection{SQA: SQL Agent}
SQA follows the serialize-everything approach: it receives a prompt containing the complete text description of the graph and the fully rendered schema (typed columns, key relations, categorical vocabularies, date ranges), totaling approximately 17\,k tokens, and produces SQL answers within at most six turns.

\begin{examplebox}{Example --- SQA on the same question}
\textbf{Trace.} The prompt already contains the founding text and every
table schema ($\approx$17\,k tokens). Turn~1: a single SQL statement
joining \texttt{tbl\_customer\_orders} and \texttt{tbl\_customers}.
Turn~2: the answer.\\[3pt]
\textbf{Outcome.} No search and no navigation, the join path is visible
in the prompt from the start. The whole corpus is read before the first
word, whether or not the question needs it, and the remaining turns serve
only to repair a failed query.
\end{examplebox}

\section{The UFK-M benchmark}

UFK-M is a fictional bicycle-assembly factory, inspired by real client
factories but entirely synthetic. A founding text states its operational
rules, KPIs and industry concepts in prose; two coordinated layers then make
that world machine-readable, a data layer (the DuckDB tables) and a semantic
layer (the knowledge graph, which distills the founding text and maps it onto
the tables). Two nested tiers scale the benchmark without removing
information; this paper evaluates the baselines mostly on \texttt{xlarge}.

\begin{table}[!htb]
\centering
\small
\renewcommand{\arraystretch}{1.3}
\begin{tabularx}{\linewidth}{l Y Y Y Y Y Y}
\toprule
\textbf{Tier} & \textbf{Tables} & \textbf{Rows} & \textbf{KG nodes} & \textbf{KG edges} & \textbf{Founding words} & \textbf{Questions} \\
\midrule
large  & 32 & 157\,953 & 125 & 268 & 4\,278 & 148 \\
\textbf{xlarge} & \textbf{64} & \textbf{174\,006} & \textbf{235} & \textbf{513} & \textbf{6\,933} & \textbf{258} \\
\bottomrule
\end{tabularx}
\caption{The two nested tiers of UFK-M.}
\label{tab:ikg-tiers}
\end{table}

\textbf{Answer-first question generation.} Questions are generated in reverse:
the answer is fixed before the question exists. For each question, an LLM
receives a sample of schema cards (table and column descriptions) and writes a SQL program over them. The program is then
executed against the database. It is kept only if its result is
non-empty and non-degenerate, and contains at most ten rows. Only then does the LLM write a natural-language question that
the retained result answers.

This order matters. Because the SQL is written and validated before the
question, every question in the set is answerable from the data, and its
gold answer is the output of a program that has actually run rather than
text produced by a model. The frozen \texttt{xlarge} set holds 258 such
questions: 116 table answers, 84 single values, 48 booleans and 10 lists;
147 need at most one join, 45 need two, 66 need three or more; 34
additionally require the semantic layer (a named operational rule or KPI
whose resolved value or formula stays hidden).

\textbf{Deterministic scoring.} Correctness is decided by a deterministic matcher rather than an LLM judge. Numeric answers are compared with tolerance for rounding precision, percentage answers are compared scale-free (independent of whether they are expressed as a fraction or as a percentage), and table answers are scored by recall of the gold rows they must contain.

\section{Experimental setup}

We evaluate seven backbone configurations across four providers: \textbf{DeepSeek~V4-Flash} (non-thinking, our reference), \textbf{DeepSeek~V4-Pro} and \textbf{V4-Pro-Think} (same weights, reasoning off vs.\ on), \textbf{GPT-5~Nano} (low/high reasoning effort), \textbf{GLM-4.5-Air}, and \textbf{Qwen3-Coder-Flash}. Each backbone runs under the three described agentic baselines, with GRA/RSA given 45 LLM-call turns and SQA 6; SQL tools cap results at 50 rows, GRA/RSA retrieve via a local embedder, \emph{multilingual-e5-large-instruct}, and SQA instead receives the full schema in its prompt with no retrieval. All models decode greedily (temperature $0$), with completions capped at 1{,}024 tokens (8{,}192 for thinking configurations) to avoid truncating answers; otherwise the harness is identical across systems. Evaluation uses the UFK-M benchmark: primary results on the validated 258-question \emph{xlarge} set. Uncertainty is quantified via paired bootstrap over questions ($B=10^4$), with 95\% percentile intervals.

\section{Results}
\label{sec:results}

\subsection{Accuracy across models}
Table~\ref{tab:accuracy-models} compares the three systems across the model grid. DeepSeek models achieve the highest accuracy for all systems, by 8--18 pp under GRA and RSA but only 2--9 pp under SQA. Among the lower-cost models, Qwen3-Coder-Flash significantly outperforms GPT-5 Nano for every system. The ranking of agentic and full-context methods depends on the model: GRA performs best with the DeepSeek and GLM models, whereas SQA performs best with Qwen3-Coder-Flash ($+2.3$\,pp over GRA) and GPT-5 Nano ($+5.5$\,pp). Full-context inference is therefore more robust when tool use is unreliable.

\begin{table}[!htb]
\centering
\small
\renewcommand{\arraystretch}{1.15}
\begin{tabularx}{\linewidth}{l Y Y Y Y}
\toprule
 & \multicolumn{3}{c}{\textbf{Accuracy (\%)}} & \\
\cmidrule(lr){2-4}
\textbf{Model} & \textbf{GRA} & \textbf{RSA} & \textbf{SQA}
  & \textbf{$\rho$ (\%)} \\
\midrule
DeepSeek V4-Flash      & \textbf{87.6} \ci{3.9} & 86.1 \ci{4.2} & 84.9 \ci{4.5}
  & \rd{$+$17.9} \\
DeepSeek V4-Pro        & \textbf{87.2} \ci{4.1} & 86.9 \ci{3.9} & 82.6 \ci{4.5}
  & \rd{$+$26.4} \\
DeepSeek V4-Pro-Think  & \textbf{88.4} \ci{3.9} & 87.4 \ci{3.9} & 83.3 \ci{4.5}
  & \rd{\textbf{$+$30.5}} \\
GLM-4.5-Air            & \textbf{77.5} \ci{5.1} & 75.6 \ci{5.2} & 74.0 \ci{5.3}
  & \rd{$+$13.5} \\
Qwen3-Coder-Flash      & 78.7 \ci{4.9} & 78.3 \ci{5.0} & \textbf{81.0} \ci{4.9}
  & \rd{$-$12.1} \\
GPT-5 Nano             & 70.5 \ci{5.9} & 70.2 \ci{5.5} & \textbf{76.0} \ci{5.3}
  & \rd{$-$22.9} \\
GPT-5 Nano (think)     & 76.7 \ci{5.1} & ---           & ---
  & \rd{---} \\
\bottomrule
\end{tabularx}
\caption{Accuracy (\%) on the frozen \texttt{xlarge} set ($n=258$), with 95\%
paired-bootstrap intervals. Bold indicates the best system for each model.
The last column reports the relative error reduction of GRA over SQA,
$\rho = (e_{\mathrm{SQA}} - e_{\mathrm{GRA}})/e_{\mathrm{SQA}}$ with
$e = 100 - \mathrm{accuracy}$; positive values mean GRA removes a fraction
of SQA's errors, negative values the reverse. Agentic methods perform best
with the DeepSeek and GLM models, whereas SQA performs best with
Qwen3-Coder-Flash and GPT-5 Nano.}
\label{tab:accuracy-models}
\end{table}

\subsection{Tool reliability matters more than extended reasoning}
Extended reasoning changes little where tool use is already reliable: the three
DeepSeek configurations lie within 1.2\,pp of one another with overlapping
intervals. GPT-5 Nano is the exception at $+6.2$\,pp, but its failure rates
roughly halve at the same time (10.2\% $\to$ 5.8\% of failed calls, 51.6\% $\to$
34.9\% of questions with failures), so the gain looks mediated by reliability rather than by
reasoning depth. Accuracy tracks that measure across the grid
(Table~\ref{tab:reliability}): the DeepSeek configurations fail under 1\% of
calls and take the top three places with the largest gains over SQA,
 GPT-5 Nano fails 10.2\% and is both the least accurate
configuration and the one for which GRA loses most to SQA. These results indicate that reliable tool use is a stronger bottleneck than extended reasoning on this task.

\begin{table}[!htb]
\centering
\small
\renewcommand{\arraystretch}{1.3}
\begin{tabularx}{\linewidth}{l Y Y Y Y}
\toprule
\textbf{Model (GRA)} & \textbf{Accuracy (\%)} & \textbf{Mean tool calls} & \textbf{Failed calls (\%)} & \textbf{Questions with $\ge 1$ failure (\%)} \\
\midrule
DeepSeek V4-Pro-Think & 88.4 & 12.1 & 0.5  & 5.4  \\
DeepSeek V4-Flash     & 87.6 & 14.8 & 0.7  & 8.1  \\
DeepSeek V4-Pro       & 87.2 & 13.4 & 0.6  & 7.0  \\
Qwen3-Coder-Flash     & 78.7 & 12.2 & 2.0  & 15.9 \\
GLM-4.5-Air           & 77.5 & 11.1 & 2.4  & 18.2 \\
GPT-5 Nano (think)    & 76.7 & 11.6 & 5.8  & 34.9 \\
GPT-5 Nano            & 70.5 & 11.1 & 10.2 & 51.6 \\
\bottomrule
\end{tabularx}
\caption{Tool-call reliability under GRA for the seven backbone configurations
of Table~\ref{tab:accuracy-models}, ordered by accuracy; the accuracy column repeats the GRA column of that table. Lower
call-failure rates coincide with higher accuracy: the three DeepSeek
configurations fail well under 1\% of calls, whereas GPT-5 Nano---the model for
which SQA outperforms GRA---fails 10.2\% and misses at least one call on more
than half of all questions. Enabling reasoning on GPT-5 Nano roughly halves
both failure measures.}
\label{tab:reliability}
\end{table}

\subsection{Token usage}
\begin{figure}[htbp]
  \centering
  \begin{tcolorbox}[colback=white, colframe=medgray,
    arc=8pt, boxrule=0.6pt,
    left=6pt, right=6pt, top=10pt, bottom=6pt,
    width=\linewidth]
    \centering
    \begin{tikzpicture}
    \begin{groupplot}[
      group style={group size=2 by 1, horizontal sep=1.5cm},
      width=7.6cm, height=5.4cm,
      ybar=1.5pt, /pgf/bar width=11pt,
      symbolic x coords={V4-Flash, V4-Pro},
      xtick=data,
      enlarge x limits=0.5,
      ymajorgrids=true, grid style={medgray!50},
      axis line style={darkgray!60},
      tick label style={font=\scriptsize, color=darkgray},
      ylabel style={font=\small, color=darkgray},
      nodes near coords,
      every node near coord/.append style={font=\tiny, color=darkgray},
    ]
    \nextgroupplot[
      ymin=0, ymax=20,
      ylabel={unique input tokens / question (k)},
      nodes near coords style={/pgf/number format/.cd, fixed, precision=1},
      legend style={at={(1.10,1.18)}, anchor=south, legend columns=3,
        draw=medgray, font=\scriptsize,
        /tikz/every even column/.append style={column sep=8pt}},
    ]
    \addplot[fill=midblue, draw=primaryblue]
      coordinates {(V4-Flash,5.6) (V4-Pro,5.0)};
    \addplot[fill=successgreen!30, draw=successgreen]
      coordinates {(V4-Flash,5.0) (V4-Pro,4.2)};
    \addplot[fill=lightorange, draw=accentorange]
      coordinates {(V4-Flash,17.2) (V4-Pro,17.2)};
    \legend{GRA, RSA, SQA}
    \nextgroupplot[
      ymin=0, ymax=2.1,
      ylabel={completion tokens / question (k)},
      nodes near coords style={/pgf/number format/.cd, fixed, precision=2},
    ]
    \addplot[fill=midblue, draw=primaryblue]
      coordinates {(V4-Flash,1.68) (V4-Pro,1.56)};
    \addplot[fill=successgreen!30, draw=successgreen]
      coordinates {(V4-Flash,1.62) (V4-Pro,1.59)};
    \addplot[fill=lightorange, draw=accentorange]
      coordinates {(V4-Flash,0.82) (V4-Pro,0.56)};
    \end{groupplot}
    \end{tikzpicture}
  \end{tcolorbox}
  \caption{Token usage per question for \textbf{DeepSeek V4-Flash} and
    \textbf{V4-Pro} (non-thinking) across the three systems. GRA and RSA denote
    the minimal-context agents; each input token is counted once per trajectory.
    \emph{Left:} the agents read approximately one-third of SQA's unique input.
    \emph{Right:} SQA produces the fewest completion tokens, while the
    multi-turn agents produce approximately two to three times as many.}
  \label{fig:tokens-deepseek}
\end{figure}
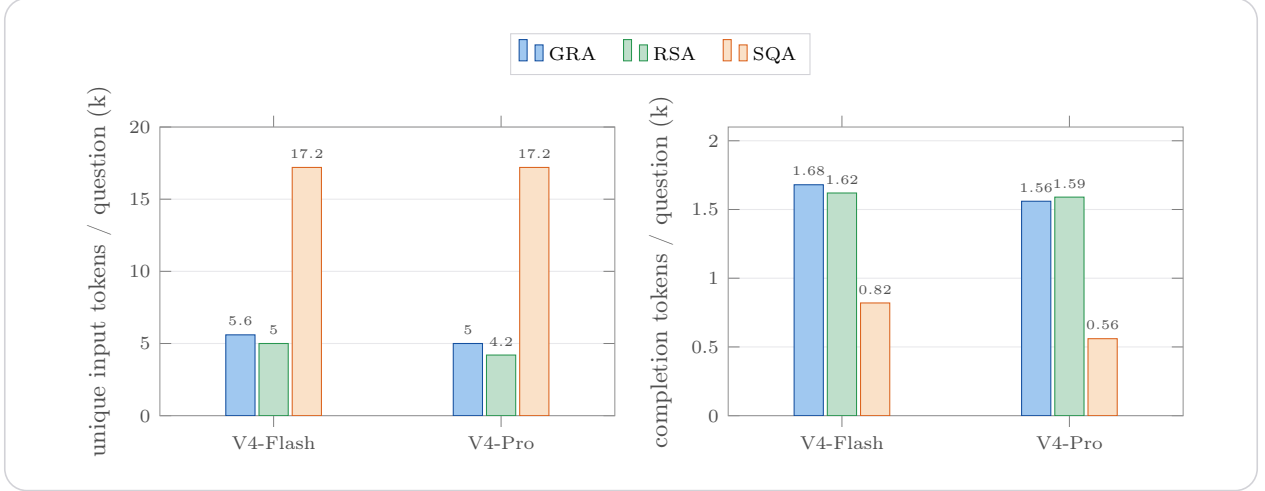

Figure~\ref{fig:tokens-deepseek} reports unique input tokens, with each token counted once per trajectory. This measures corpus coverage rather than billed usage. Across the five models, GRA reads 29--33\% of SQA's unique input, while RSA reads 24--29\%. Graph navigation therefore reads slightly more input than flat textual retrieval, but both agents remain well below the full-context baseline. The pattern reverses for output tokens. SQA produces 0.5--1.3\,k completion tokens per question because it answers with a single SQL block, whereas the multi-turn agents produce 1.0--2.0\,k during iterative search and planning.

Unique and billed input tokens differ. GRA and RSA agents resend a growing context over 11--15 turns, whereas SQA averages fewer than three turns. Under cache-aware pricing, the relative cost depends on the workload. Warm-cache batch evaluation favours SQA because its large fixed prefix can be reused across questions. Cold-start single-question serving favours GRA and RSA agents because SQA incurs its full 17\,k-token prompt on every question.

\subsection{Effect of the tool-call budget}

Each question is subject to a fixed tool-call budget. If the budget is exhausted before the relevant data are found, the agent must answer before completing its search. We vary the budget $B$ over $\{10, 20, 30, 40, 50\}$ on the \emph{large} ($N=148$) and \emph{xlarge} ($N=258$) sets. We use the budget-aware \texttt{GRA} agent, which receives the budget at the start and a warning after approximately 80\% has been consumed. In addition to accuracy, we measure the \emph{truncation rate}, defined as the proportion of questions that exhaust the budget without a voluntary \texttt{answer}, and the mean number of tool-calling turns used.
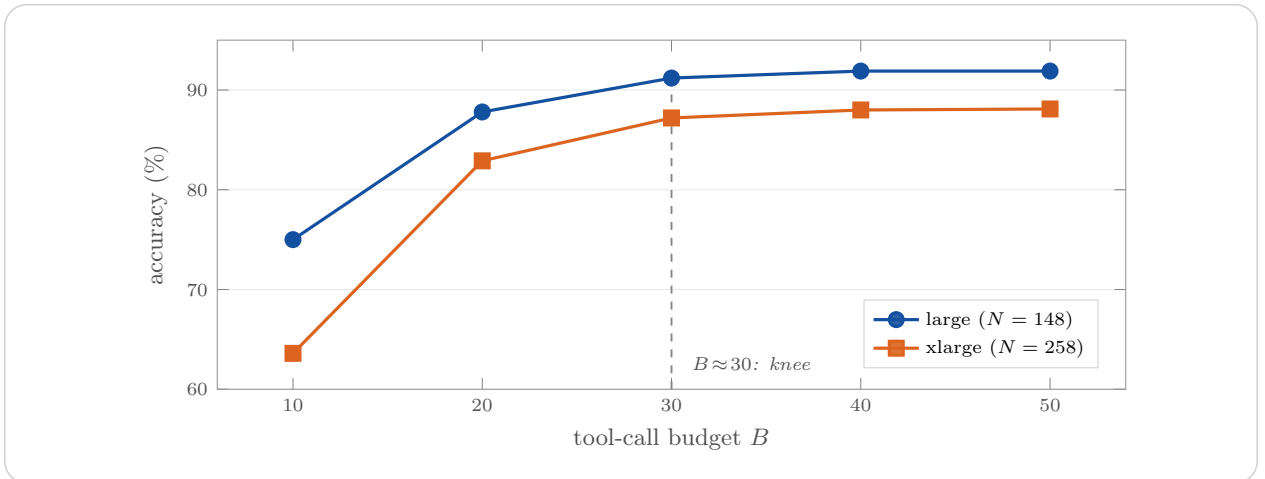
\begin{figure}[htbp]
  \centering
  \begin{tcolorbox}[colback=white, colframe=medgray,
    arc=8pt, boxrule=0.6pt,
    left=6pt, right=6pt, top=10pt, bottom=6pt,
    width=\linewidth]
    \centering
    \begin{tikzpicture}
    \begin{axis}[
      width=13.6cm, height=6.2cm,
      xlabel={tool-call budget $B$},
      ylabel={accuracy (\%)},
      xlabel style={font=\small, color=darkgray},
      ylabel style={font=\small, color=darkgray},
      xmin=6, xmax=54,
      ymin=60, ymax=95,
      xtick={10,20,30,40,50},
      ytick={60,70,80,90},
      ymajorgrids=true, grid style={medgray!50},
      axis line style={darkgray!60},
      tick label style={font=\scriptsize, color=darkgray},
      legend style={at={(0.97,0.06)}, anchor=south east, draw=medgray,
                    font=\scriptsize, fill=white, fill opacity=0.9, text opacity=1},
      legend cell align={left},
      every axis plot/.append style={line width=1.2pt, mark size=2.6pt},
      clip=false,
    ]
    \addplot[color=primaryblue, mark=*, mark options={fill=primaryblue}] coordinates
      {(10,75.0) (20,87.8) (30,91.2) (40,91.9) (50,91.9)};
    \addlegendentry{large ($N=148$)}
    \addplot[color=accentorange, mark=square*, mark options={fill=accentorange}] coordinates
      {(10,63.6) (20,82.9) (30,87.2) (40,88.0) (50,88.1)};
    \addlegendentry{xlarge ($N=258$)}
    \draw[dashed, darkgray!70, line width=0.7pt] (axis cs:30,60) -- (axis cs:30,91.2);
    \node[font=\scriptsize\itshape, color=darkgray, anchor=south west]
      at (axis cs:30.6,61) {$B\!\approx\!30$: knee};
    \end{axis}
    \end{tikzpicture}
  \end{tcolorbox}
  \caption{Accuracy as a function of the tool-call budget $B$ for \texttt{GRA}
    with DeepSeek V4-Flash. Accuracy increases sharply from $B=10$ to $B=30$ and
    then plateaus on both datasets. A budget of approximately 30 calls captures
    nearly all observed gains.}
  \label{fig:budget}
\end{figure}

Accuracy increases sharply between 10 and 30 calls and then plateaus. Over the same range, the number of truncated questions decreases from 62 to 3 on \emph{large} and from 118 to 11 on \emph{xlarge}. Once truncation becomes rare, the mean number of turns stabilizes at approximately 11 on \emph{large} and 13 on \emph{xlarge}, and additional budget has little effect. Budgets below 20 calls frequently truncate the search and reduce accuracy; on \emph{xlarge}, accuracy is 63.6\% at $B=10$. No measurable gain is observed above 30 calls, making a budget of approximately 30 calls sufficient for these datasets.

\section{Further analysis}
\label{sec:analysis}

Taken together, the results identify selective agentic access as the primary source of GRA's advantage on this benchmark. On models that call tools reliably, both agents outperform SQA, and GRA achieves the largest well-supported gain at $+5.1$\,pp while reading only 29--33\% of SQA's unique input tokens. This efficiency is expected to matter more as the corpus grows beyond what a single prompt can hold, since SQA's cost scales with the full serialized context whereas the agents read only the fraction they retrieve. The contribution of graph topology, however, is much less clear: GRA exceeds RSA by only $+0.3$\, to $+1.9$\,pp, and the two are indistinguishable on DeepSeek V4-Pro, indicating that the gain over SQA comes mainly from how context is accessed rather than from graph structure itself. Reliability sets the boundary of this behaviour, as the advantage of the agentic harness shrinks or reverses once tool-call failures exceed a few percent, which is why full-context inference remains preferable for weaker tool-callers. Finally, the current corpus is small enough that SQA's 17\,k-token prompt fits comfortably in every model's context window, so the regime in which structured navigation should help most, where serialization is infeasible or too costly, remains to be tested on substantially larger graphs.

\section{GRA brings industrial intelligence --- complete overview}
\label{sec:overview}

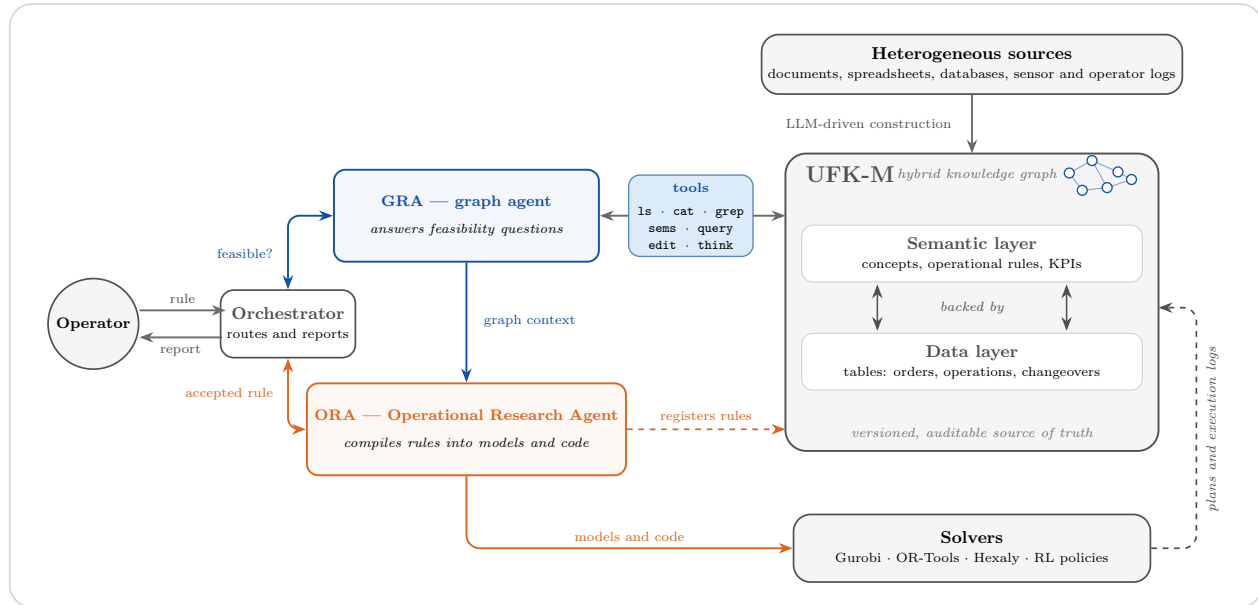
\begin{figure}[htbp]
  \centering
  \begin{tcolorbox}[colback=white, colframe=medgray, arc=8pt, boxrule=0.6pt,
    left=4pt, right=4pt, top=8pt, bottom=6pt, width=\linewidth]
    \begin{center}
    \resizebox{0.98\linewidth}{!}{
    \begin{tikzpicture}[
      font=\small,
      >={Stealth[length=2.6mm]},
      agentbox/.style={rounded corners=6pt, draw=#1, line width=1pt,
                       fill=#1!5, align=center, inner sep=4pt},
      graybox/.style={draw=darkgray, line width=0.9pt, fill=lightgray,
                      rounded corners=9pt, align=center},
      innerwhite/.style={draw=medgray, line width=0.6pt, fill=white,
                         rounded corners=6pt, align=center},
      flow/.style={line width=1.0pt, darkgray!85},
      kgnode/.style={circle, draw=primaryblue, fill=white, line width=0.7pt,
                     minimum size=5.2pt, inner sep=0pt}
  ]

  \draw[draw=darkgray, line width=1.0pt, fill=lightgray, rounded corners=11pt]
      (10.9,-4.0) rectangle (17.8,1.5);
  \node[anchor=west, font=\large\bfseries, text=darkgray] at (11.15,1.12) {UFK-M};
  \node[anchor=west, font=\scriptsize\itshape, text=darkgray!85] at (12.85,1.10)
      {hybrid knowledge graph};

  \begin{scope}[shift={(16.15,0.82)}]
      \draw[primaryblue!55, line width=0.55pt]
          (0,0.32) -- (0.42,0.55) -- (0.90,0.34)
          (0,0.32) -- (0.24,0.0) -- (0.70,0.05) -- (1.15,0.20)
          (0.42,0.55) -- (0.70,0.05)
          (0.90,0.34) -- (1.15,0.20);
      \node[kgnode] at (0,0.32) {};   \node[kgnode] at (0.42,0.55) {};
      \node[kgnode] at (0.90,0.34) {};\node[kgnode] at (0.24,0.0) {};
      \node[kgnode] at (0.70,0.05) {};\node[kgnode] at (1.15,0.20) {};
  \end{scope}

  \node[innerwhite, minimum width=6.3cm, minimum height=1.05cm] (semL) at (14.35,-0.35)
      {\textbf{\textcolor{darkgray}{Semantic layer}}\\[-1pt]
       \scriptsize concepts, operational rules, KPIs};
  \node[innerwhite, minimum width=6.3cm, minimum height=1.05cm] (datL) at (14.35,-2.35)
      {\textbf{\textcolor{darkgray}{Data layer}}\\[-1pt]
       \scriptsize tables: orders, operations, changeovers};
  \draw[<->, darkgray, line width=0.8pt] (12.6,-0.92) -- (12.6,-1.78);
  \draw[<->, darkgray, line width=0.8pt] (16.1,-0.92) -- (16.1,-1.78);
  \node[font=\scriptsize\itshape, text=darkgray] at (14.35,-1.35) {backed by};
  \node[font=\scriptsize\itshape, text=darkgray!80] at (14.35,-3.68)
      {versioned, auditable source of truth};

  \node[graybox, minimum width=7.1cm, minimum height=1.1cm] (src) at (14.35,3.15)
      {\textbf{Heterogeneous sources}\\[-1pt]
       \scriptsize documents, spreadsheets, databases, sensor and operator logs};
  \draw[->, flow] (src.south) -- (14.35,1.5);
  \node[anchor=east, align=right, font=\scriptsize, text=darkgray] at (14.1,2.05)
      {LLM-driven construction};

  \node[agentbox=primaryblue, minimum width=4.9cm, minimum height=1.7cm] (gra) at (5.0,0.35)
      {{\footnotesize\bfseries\color{primaryblue}GRA --- graph agent}\\[3pt]
       {\scriptsize\itshape answers feasibility questions}};

  \draw[<->, flow] (gra.east) -- (10.9,0.35);
  \node[draw=primaryblue!70, fill=lightblue, rounded corners=6pt, line width=0.7pt,
        align=center, inner sep=4pt, minimum width=2.3cm] (tools) at (9.15,0.35)
      {{\scriptsize\bfseries\color{primaryblue}tools}\\[1pt]
       \texttt{\scriptsize ls · cat · grep}\\[-2pt]
       \texttt{\scriptsize sems · query}\\[-2pt]
       \texttt{\scriptsize edit · think}};

  \node[agentbox=accentorange, minimum width=4.9cm, minimum height=1.7cm] (ora) at (5.0,-3.6)
      {{\footnotesize\bfseries\color{accentorange}ORA --- Operational Research Agent}\\[3pt]
       {\scriptsize\itshape compiles rules into models and code}};

  \draw[->, primaryblue, line width=1.0pt] (gra.south) -- (ora.north);
  \node[anchor=west, align=left, font=\scriptsize, text=primaryblue] at (5.2,-1.6)
      {graph context};

  \node[draw=darkgray, fill=white, line width=0.9pt, rounded corners=7pt,
        align=center, minimum width=2.4cm, minimum height=1.25cm] (orch) at (1.7,-1.65)
      {\textbf{\color{darkgray}Orchestrator}\\[-1pt]
       \scriptsize routes and reports};

  \node[circle, draw=darkgray, fill=lightgray, line width=0.9pt, minimum size=1.35cm,
        align=center, font=\footnotesize\bfseries] (user) at (-1.9,-1.65) {Operator};

  \draw[->, flow] (user.east|-0.01,-1.4) -- (0.55,-1.4)
      node[midway, above, font=\scriptsize, text=darkgray] {rule};
  \draw[->, flow] (0.48,-1.9) -- (user.east|-0,-1.9)
      node[midway, below, font=\scriptsize, text=darkgray] {report};

  \draw[<->, primaryblue, line width=1.0pt, rounded corners=6pt]
      (orch.north) |- (gra.west);
  \node[anchor=east, align=right, font=\scriptsize, text=primaryblue] at (1.55,-0.35)
      {feasible?};

  \draw[<->, accentorange, line width=1.0pt, rounded corners=6pt]
      (orch.south) |- (ora.west);
  \node[anchor=east, align=right, font=\scriptsize, text=accentorange] at (1.55,-2.95)
      {accepted rule};

  \draw[->, accentorange, dashed, line width=0.9pt] (ora.east) -- (10.9,-3.6)
      node[midway, above, font=\scriptsize, text=accentorange] {registers rules};

  \node[graybox, minimum width=6.6cm, minimum height=1.25cm] (solv) at (14.35,-5.8)
      {\textbf{Solvers}\\[-1pt]
       \scriptsize Gurobi · OR-Tools · Hexaly · RL policies};
  \draw[->, accentorange, line width=1.0pt, rounded corners=7pt]
      (ora.south) |- (solv.west)
      node[pos=0.75, above, font=\scriptsize, text=accentorange]
      {models and code};

  \draw[->, dashed, darkgray, line width=0.9pt, rounded corners=7pt]
      (solv.east) -- (18.5,-5.8) -- (18.5,-1.35) -- (17.8,-1.35);
  \node[rotate=90, font=\scriptsize\itshape, text=darkgray] at (18.8,-3.6)
      {plans and execution logs};

  \end{tikzpicture}
    }
    \end{center}
  \end{tcolorbox}
  \caption{Deployment loop around the benchmarked block. The operator states a
    rule in plain language; the orchestrator asks GRA for a feasibility
    verdict, which GRA grounds by navigating the hybrid graph with the generic
    tools of Table~\ref{tab:tools}. Accepted rules go to ORA, which compiles
    them into mathematical optimization models and solver code, and registers
    them back into the graph. Plans and execution logs return to the data
    layer, so the substrate accumulates what the loop decides. GRA (blue) is
    the block evaluated in Sections~\ref{sec:results}--\ref{sec:analysis}; the
    rest is the deployment setting around it.}
  \label{fig:deploy}
\end{figure}

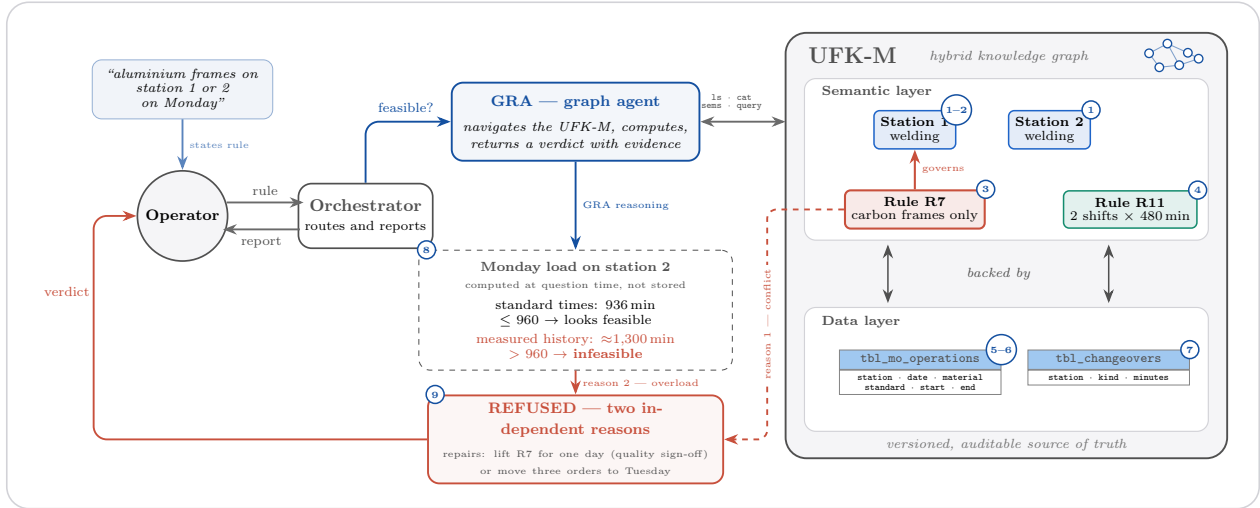
\begin{figure}[htbp]
  \centering
  \begin{tcolorbox}[colback=white, colframe=medgray, arc=8pt, boxrule=0.6pt,
    left=4pt, right=4pt, top=8pt, bottom=6pt, width=\linewidth]
    \begin{center}
    \resizebox{0.98\linewidth}{!}{%
    \begin{tikzpicture}[
        font=\small,
        >={Stealth[length=2.6mm]},
        agentbox/.style={rounded corners=6pt, draw=#1, line width=1pt,
                         fill=#1!5, align=center, inner sep=4pt},
        chip/.style={rounded corners=3pt, draw=#1, fill=#1!12, thick,
                     align=center, font=\scriptsize, inner sep=3.5pt},
        tblh/.style={draw=darkgray!70, fill=midblue, minimum width=3.0cm,
                     minimum height=0.36cm, inner sep=1.5pt,
                     font=\scriptsize\ttfamily\bfseries, text=darkgray},
        tblb/.style={draw=darkgray!70, fill=white, minimum width=3.0cm,
                     inner sep=2.4pt, font=\tiny\ttfamily, align=center},
        derived/.style={draw=darkgray, dashed, fill=white, rounded corners=6pt,
                        align=center, font=\scriptsize, inner sep=6pt,
                        text width=5.4cm},
        badge/.style={circle, draw=primaryblue, fill=white, text=primaryblue,
                      font=\tiny\bfseries, inner sep=1.6pt, line width=0.8pt},
        flow/.style={line width=1.0pt, darkgray!85},
        kgnode/.style={circle, draw=primaryblue, fill=white, line width=0.7pt,
                       minimum size=4.5pt, inner sep=0pt}
    ]
 
    \draw[draw=darkgray, line width=1.0pt, fill=lightgray, rounded corners=11pt]
        (9.6,-3.7) rectangle (17.8,4.2);
    \node[anchor=west, font=\large\bfseries, text=darkgray] at (9.9,3.78) {UFK-M};
    \node[anchor=west, font=\scriptsize\itshape, text=darkgray!85] at (12.15,3.75)
        {hybrid knowledge graph};
    \begin{scope}[shift={(16.35,3.56)}]
        \draw[primaryblue!55, line width=0.55pt]
            (0,0.26) -- (0.34,0.44) -- (0.72,0.27)
            (0,0.26) -- (0.19,0) -- (0.56,0.04) -- (0.92,0.16)
            (0.34,0.44) -- (0.56,0.04) (0.72,0.27) -- (0.92,0.16);
        \node[kgnode] at (0,0.26) {};   \node[kgnode] at (0.34,0.44) {};
        \node[kgnode] at (0.72,0.27) {};\node[kgnode] at (0.19,0) {};
        \node[kgnode] at (0.56,0.04) {};\node[kgnode] at (0.92,0.16) {};
    \end{scope}
 
    \draw[draw=medgray, line width=0.6pt, fill=white, rounded corners=6pt]
        (9.95,0.35) rectangle (17.45,3.35);
    \node[anchor=west, font=\scriptsize\bfseries, text=darkgray] at (10.15,3.11)
        {Semantic layer};
 
    \node[chip=nodeblue] (st1) at (12.0,2.40) {\textbf{Station 1}\\[-1pt] welding};
    \node[chip=nodeblue] (st2) at (14.5,2.40) {\textbf{Station 2}\\[-1pt] welding};
    \node[chip=nodecoral, line width=1.1pt] (r7) at (12.0,0.92)
        {\textbf{Rule R7}\\[-1pt] carbon frames only};
    \node[chip=nodeteal] (r11) at (16.0,0.92)
        {\textbf{Rule R11}\\[-1pt] 2 shifts $\times$ 480\,min};
 
    \draw[->, nodecoral, line width=1.1pt] (r7.north) -- (st1.south)
        node[midway, anchor=west, font=\tiny, text=nodecoral] {governs};
 
    \draw[draw=medgray, line width=0.6pt, fill=white, rounded corners=6pt]
        (9.95,-3.2) rectangle (17.45,-0.9);
    \node[anchor=west, font=\scriptsize\bfseries, text=darkgray] at (10.15,-1.17)
        {Data layer};
 
    \node[tblh] (t1h) at (12.1,-1.87) {tbl\_mo\_operations};
    \node[tblb, anchor=north] (t1b) at (t1h.south)
        {station · date · material\\[-1pt] standard · start · end};
    \node[tblh] (t2h) at (15.6,-1.87) {tbl\_changeovers};
    \node[tblb, anchor=north] (t2b) at (t2h.south)
        {station · kind · minutes};
 
    \draw[<->, darkgray, line width=0.8pt] (11.5,0.27) -- (11.5,-0.82);
    \draw[<->, darkgray, line width=0.8pt] (15.6,0.27) -- (15.6,-0.82);
    \node[font=\scriptsize\itshape, text=darkgray] at (13.55,-0.28) {backed by};
    \node[font=\scriptsize\itshape, text=darkgray!80] at (13.7,-3.45)
        {versioned, auditable source of truth};
 
    \node[draw=primaryblue!45, fill=primaryblue!6, rounded corners=5pt,
          align=center, font=\scriptsize\itshape, inner sep=5pt,
          text width=3.0cm] (quote) at (-1.6,3.15)
        {``aluminium frames on\\[-1pt] station 1 or 2 on Monday''};
 
    \node[circle, draw=darkgray, fill=lightgray, line width=0.9pt,
          minimum size=1.35cm, align=center, font=\footnotesize\bfseries]
          (user) at (-1.6,0.8) {Operator};
    \node[draw=darkgray, fill=white, line width=0.9pt, rounded corners=7pt,
          align=center, minimum width=2.3cm, minimum height=1.2cm]
          (orch) at (1.8,0.8)
        {\textbf{\color{darkgray}Orchestrator}\\[-1pt]
         \scriptsize routes and reports};
 
    \draw[->, primaryblue!70, line width=1.0pt] (quote.south) -- (user.north)
        node[midway, anchor=west, font=\tiny, text=primaryblue!70] {states rule};
 
    \draw[->, flow] (-0.78,1.05) -- (0.65,1.05)
        node[midway, above, font=\scriptsize, text=darkgray] {rule};
    \draw[->, flow] (0.58,0.55) -- (-0.85,0.55)
        node[midway, below, font=\scriptsize, text=darkgray] {report};
 
    \node[agentbox=primaryblue, minimum width=4.6cm, minimum height=1.45cm]
          (gra) at (5.7,2.55)
        {{\footnotesize\bfseries\color{primaryblue}GRA --- graph agent}\\[2pt]
         {\scriptsize\itshape navigates the UFK-M, computes,}\\[-2pt]
         {\scriptsize\itshape returns a verdict with evidence}};
 
    \draw[->, primaryblue, line width=1.0pt, rounded corners=7pt]
        (orch.north) |- (gra.west);
    \node[anchor=south, font=\scriptsize, text=primaryblue] at (2.55,2.61)
        {feasible?};
 
    \draw[<->, flow] (gra.east) -- (9.6,2.55);
    \node[anchor=south, align=center, font=\tiny\ttfamily, text=darkgray]
        at (8.6,2.65) {ls · cat\\[-2pt] sems · query};
 
    \node[derived] (der) at (5.7,-0.95)
        {\textbf{\color{darkgray}Monday load on station 2}\\[1pt]
         {\tiny\upshape\color{darkgray} computed at question time, not stored}\\[3pt]
         standard times: 936\,min $\le$ 960 $\to$ looks feasible\\[2pt]
         \textcolor{nodecoral}{measured history:
         $\approx$1{,}300\,min $>$ 960 $\to$ \textbf{infeasible}}};
    \draw[->, primaryblue, line width=1.0pt] (gra.south) -- (der.north)
        node[midway, anchor=west, font=\tiny, text=primaryblue]
        {GRA reasoning};
 
    \node[agentbox=nodecoral, text width=5.2cm] (ver) at (5.7,-3.35)
        {{\footnotesize\bfseries\color{nodecoral}REFUSED --- two independent reasons}\\[1pt]
         {\tiny\color{darkgray} repairs: lift R7 for one day (quality sign-off)}\\[-2pt]
         {\tiny\color{darkgray} or move three orders to Tuesday}};
         
    \draw[->, nodecoral, line width=1.0pt] (der.south) -- (ver.north)
        node[midway, anchor=west, font=\tiny, text=nodecoral]
        {reason 2 --- overload};
 
    \coordinate (rail) at ($(ver.east)+(0.8,0)$);
    \draw[->, nodecoral, dashed, line width=1.0pt, rounded corners=5pt]
        (r7.west) -- (r7.west -| rail) -- (rail) -- (ver.east);
    \node[rotate=90, fill=white, inner sep=1.5pt, font=\tiny, text=nodecoral]
        at ($(rail)+(0,2.15)$) {reason 1 --- conflict};
 
    \draw[->, nodecoral, line width=1.0pt, rounded corners=7pt]
        (ver.west) -- (-3.25,-3.35) -- (-3.25,0.8) -- (user.west);
    \node[anchor=east, font=\scriptsize, text=nodecoral] at (-3.2,-0.6)
        {verdict};
 
    \node[badge] at (st1.north east) {1--2};
    \node[badge] at (st2.north east) {1};
    \node[badge] at (r7.north east)  {3};
    \node[badge] at (r11.north east) {4};
    \node[badge] at (t1h.north east) {5--6};
    \node[badge] at (t2h.north east) {7};
    \node[badge] at ($(der.north west)+(0.15,0)$) {8};
    \node[badge] at ($(ver.north west)+(0.15,0)$) {9};
 
    \end{tikzpicture}%
    }
    \end{center}
  \end{tcolorbox}
  \caption{Example~1 on the architecture of Figure~\ref{fig:deploy}. An operator
    states a rule in plain language; it enters through the orchestrator, and GRA
    explores the UFK-M, whose semantic layer holds the concepts and the rules
    and whose data layer holds the tables (badges 1--9 give the call order of
    the tool-call trace). Two findings force the refusal: rule R7 governs
    station~1, and the dashed box, computed at question time from the two tables
    and the calendar of R11, shows that station~2 alone cannot absorb the daily
    workload.}
  \label{fig:example-refuse}
\end{figure}

\begin{figure}[htbp]
  \centering
  \begin{tcolorbox}[colback=white, colframe=medgray, arc=8pt, boxrule=0.6pt,
    left=4pt, right=4pt, top=8pt, bottom=6pt, width=\linewidth]
    \begin{center}
    \resizebox{0.98\linewidth}{!}{%
    \begin{tikzpicture}[
        font=\small,
        >={Stealth[length=2.6mm]},
        agentbox/.style={rounded corners=6pt, draw=#1, line width=1pt,
                         fill=#1!5, align=center, inner sep=4pt},
        chip/.style={rounded corners=3pt, draw=#1, fill=#1!12, thick,
                     align=center, font=\scriptsize, inner sep=3.5pt},
        tblh/.style={draw=darkgray!70, fill=midblue, minimum width=3.0cm,
                     minimum height=0.36cm, inner sep=1.5pt,
                     font=\scriptsize\ttfamily\bfseries, text=darkgray},
        tblb/.style={draw=darkgray!70, fill=white, minimum width=3.0cm,
                     inner sep=2.4pt, font=\tiny\ttfamily, align=center},
        mathbox/.style={draw=accentorange!60, fill=accentorange!7,
                        rounded corners=4pt, align=center, inner sep=4pt,
                        minimum width=3.4cm},
        codeh/.style={draw=darkgray!70, fill=medgray!45, minimum width=3.4cm,
                      minimum height=0.4cm, inner sep=1.5pt,
                      font=\scriptsize\ttfamily\bfseries, text=darkgray},
        codeb/.style={draw=darkgray!70, minimum width=3.4cm, inner sep=2.4pt,
                      font=\tiny\ttfamily, align=center},
        derived/.style={draw=darkgray, dashed, fill=white, rounded corners=6pt,
                        align=center, font=\scriptsize, inner sep=6pt,
                        text width=5.4cm},
        badge/.style={circle, draw=primaryblue, fill=white, text=primaryblue,
                      font=\tiny\bfseries, inner sep=1.6pt, line width=0.8pt},
        flow/.style={line width=1.0pt, darkgray!85},
        kgnode/.style={circle, draw=primaryblue, fill=white, line width=0.7pt,
                       minimum size=4.5pt, inner sep=0pt},
        merge/.style={circle, fill=successgreen, minimum size=3.2pt, inner sep=0pt}
    ]
 
    \draw[draw=darkgray, line width=1.0pt, fill=lightgray, rounded corners=11pt]
        (9.6,-3.05) rectangle (17.8,3.8);
    \node[anchor=west, font=\large\bfseries, text=darkgray] at (9.9,3.38) {UFK-M};
    \node[anchor=west, font=\scriptsize\itshape, text=darkgray!85] at (12.15,3.35)
        {hybrid knowledge graph};
    \begin{scope}[shift={(16.35,3.16)}]
        \draw[primaryblue!55, line width=0.55pt]
            (0,0.26) -- (0.34,0.44) -- (0.72,0.27)
            (0,0.26) -- (0.19,0) -- (0.56,0.04) -- (0.92,0.16)
            (0.34,0.44) -- (0.56,0.04) (0.72,0.27) -- (0.92,0.16);
        \node[kgnode] at (0,0.26) {};   \node[kgnode] at (0.34,0.44) {};
        \node[kgnode] at (0.72,0.27) {};\node[kgnode] at (0.19,0) {};
        \node[kgnode] at (0.56,0.04) {};\node[kgnode] at (0.92,0.16) {};
    \end{scope}
 
    \draw[draw=medgray, line width=0.6pt, fill=white, rounded corners=6pt]
        (9.95,0.55) rectangle (17.45,3.0);
    \node[anchor=west, font=\scriptsize\bfseries, text=darkgray] at (10.15,2.76)
        {Semantic layer};
 
    \node[chip=nodepurple] (cc)  at (11.5,2.08)
        {\textbf{Concept}\\[-1pt] colour change};
    \node[chip=nodeteal]   (r11) at (13.9,2.08)
        {\textbf{Rule R11}\\[-1pt] 2 shifts $\times$ 480\,min};
    \node[chip=nodeblue]   (l1)  at (16.2,2.08)
        {\textbf{Line 1}\\[-1pt] assembly};
    \node[chip=successgreen, dashed] (r23) at (16.2,1.05)
        {\textbf{Rule R23 (new)}\\[-1pt] $\le 3$ changes/shift};
 
    \draw[->, successgreen, line width=1.0pt] (r23.north) -- (l1.south)
        node[midway, anchor=west, font=\tiny, text=successgreen] {constrains};
 
    \draw[draw=medgray, line width=0.6pt, fill=white, rounded corners=6pt]
        (9.95,-2.6) rectangle (17.45,-0.35);
    \node[anchor=west, font=\scriptsize\bfseries, text=darkgray] at (10.15,-0.62)
        {Data layer};
 
    \node[tblh] (t2h) at (12.1,-1.32) {tbl\_changeovers};
    \node[tblb, anchor=north] (t2b) at (t2h.south)
        {station · kind · minutes};
    \node[tblh] (t1h) at (15.6,-1.32) {tbl\_mo\_operations};
    \node[tblb, anchor=north] (t1b) at (t1h.south)
        {line · date · shift · colour};
 
    \draw[<->, darkgray, line width=0.8pt] (11.5,0.47) -- (11.5,-0.27);
    \draw[<->, darkgray, line width=0.8pt] (15.6,0.47) -- (15.6,-0.27);
    \node[font=\scriptsize\itshape, text=darkgray] at (13.55,0.1) {backed by};
    \node[font=\scriptsize\itshape, text=darkgray!80] at (13.7,-2.82)
        {versioned, auditable source of truth};
 
    \node[draw=primaryblue!45, fill=primaryblue!6, rounded corners=5pt,
          align=center, font=\scriptsize\itshape, inner sep=5pt,
          text width=3.2cm] (quote) at (-1.6,2.85)
        {``at most three colour changes\\[-1pt] per shift on line 1''};
 
    \node[circle, draw=darkgray, fill=lightgray, line width=0.9pt,
          minimum size=1.35cm, align=center, font=\footnotesize\bfseries]
          (user) at (-1.6,0.7) {Operator};
    \node[draw=darkgray, fill=white, line width=0.9pt, rounded corners=7pt,
          align=center, minimum width=2.3cm, minimum height=1.2cm]
          (orch) at (1.6,0.7)
        {\textbf{\color{darkgray}Orchestrator}\\[-1pt]
         \scriptsize routes \& reports};
 
    \draw[->, primaryblue!70, line width=1.0pt] (quote.south) -- (user.north)
        node[midway, anchor=west, font=\tiny, text=primaryblue!70] {states rule};
 
    \draw[->, flow] (-0.65,0.95) -- (0.45,0.95)
        node[midway, above, font=\scriptsize, text=darkgray] {rule};
    \draw[->, flow] (0.45,0.45) -- (-0.65,0.45)
        node[midway, below, font=\scriptsize, text=darkgray] {report};
 
    \node[agentbox=primaryblue, minimum width=4.6cm, minimum height=1.45cm]
          (gra) at (5.55,2.1)
        {{\footnotesize\bfseries\color{primaryblue}GRA --- graph agent}\\[2pt]
         {\scriptsize\itshape navigates the UFK-M, computes,}\\[-2pt]
         {\scriptsize\itshape returns a verdict with evidence}};
 
    \draw[->, primaryblue, line width=1.0pt, rounded corners=7pt]
        (orch.north) |- (gra.west);
    \node[anchor=south, font=\scriptsize, text=primaryblue] at (2.35,2.16)
        {feasible?};
 
    \draw[<->, flow] (gra.east) -- (9.6,2.1);
    \node[anchor=south, align=center, font=\tiny\ttfamily, text=darkgray]
        at (8.6,2.2) {ls · cat\\[-2pt] sems · query};
 
    \node[derived, fill=successgreen!8, draw=successgreen] (der) at (5.55,-1.35)
        {\textbf{\color{successgreen}FEASIBLE --- risk measured}\\[2pt]
         {\upshape\color{darkgray} Replay of 428 shifts,}\\[-1pt]
         {\tiny\upshape\color{darkgray} computed at question time, not stored}\\[3pt]
         \textcolor{nodecoral}{11 shifts break the cap}};
    \draw[->, primaryblue, line width=1.0pt] (gra.south) -- (der.north)
        node[midway, anchor=west, font=\tiny, text=primaryblue]
        {GRA reasoning};
 
    \draw[->, successgreen, line width=1.0pt, rounded corners=7pt]
        (der.west) -- (-3.05,-1.35) -- (-3.05,0.7) -- (user.west);
    \node[anchor=south, font=\scriptsize, text=successgreen] at (-2.5,-0.55)
        {verdict};
 
    \node[agentbox=accentorange, minimum width=4.6cm, minimum height=1.35cm]
          (ora) at (5.55,-4.7)
        {{\footnotesize\bfseries\color{accentorange}ORA --- Operational Research Agent}\\[2pt]
         {\scriptsize\itshape compiles the accepted rule}};
    \draw[->, accentorange, line width=1.0pt, rounded corners=7pt]
        (orch.south) -- (1.6,-4.7) -- (ora.west);
    \node[anchor=east, align=right, font=\tiny, text=accentorange]
        at (1.15,-3.1)
        {operator confirms /\\[-1pt] orchestrator routes};
 
    \node[mathbox] (math) at (10.9,-4.05)
        {{\tiny\bfseries\color{accentorange}mathematical model}\\[2pt]
         {\footnotesize $\displaystyle\sum_{s}\mathrm{chg}_s \le 3$}\\[1pt]
         {\tiny\color{darkgray} per shift, with $\mathrm{chg}_s\!\ge\!k_s\!-\!1$}};
 
    \node[codeh] (mh) at (10.9,-5.55) {scheduling.mzn};
    \node[codeb, fill=white, anchor=north] (mb) at (mh.south)
        {\% existing model \dots};
    \node[codeb, fill=successgreen!14, draw=successgreen, anchor=north] (ma) at (mb.south)
        {+ seq[$\cdot$], change[$\cdot$]\\[-1pt]
         + cap $\le 3$ per shift\\[-1pt]
         + floor $\ge k-1$};
 
    \draw[->, accentorange, line width=1.0pt, rounded corners=6pt]
        (ora.east) -- (8.8,-4.7) -- (8.8,-4.05) -- (math.west);
    \draw[->, accentorange, line width=1.0pt, rounded corners=6pt]
        (ora.east) -- (8.8,-4.7) -- (8.8,-5.55) -- (mh.west);
 
    \coordinate (mjoin) at (13.7,-4.8);
    \draw[successgreen, line width=0.9pt, rounded corners=6pt]
        (math.east) -| (mjoin);
    \draw[successgreen, line width=0.9pt, rounded corners=6pt]
        (mh.east) -| (mjoin);
    \node[merge] at (mjoin) {};
    \draw[->, successgreen, dashed, line width=0.9pt, rounded corners=7pt]
        (mjoin) -- (18.2,-4.8) -- (18.2,1.05) -- (r23.east);
    \node[rotate=90, font=\scriptsize\itshape, text=successgreen]
        at (18.5,-1.9) {registers R23 (\texttt{edit})};
 
    \node[badge] at (cc.north east)  {1};
    \node[badge] at (t2h.north east) {1};
    \node[badge] at (r11.north east) {2};
    \node[badge] at (l1.north east)  {3};
    \node[badge] at (t1h.north east) {4};
    \node[badge] at ($(der.north west)+(0.15,0)$) {5--6};
 
    \end{tikzpicture}%
    }
    \end{center}
  \end{tcolorbox}
  \caption{Example~2 on the architecture of Figure~\ref{fig:deploy}. An operator
    states a rule in plain language; GRA resolves the words through the semantic
    layer (badges 1--4) and replays ten months of history from the data layer,
    returning a single feasible verdict with a measured risk (badge 5--6). The
    verdict goes back to the operator; once confirmed, the orchestrator routes
    the accepted rule to ORA. ORA produces it in two forms --- a mathematical
    optimization model and the corresponding \texttt{scheduling.mzn} solver code
    (new variables, new constraint) --- whose outputs merge and register the
    rule back into the semantic layer as node R23 (right-side dashed loop, as in
    Figure~\ref{fig:deploy}), where it now constrains Line~1. The graph
    accumulates the decision.}
  \label{fig:example-compile}
\end{figure}
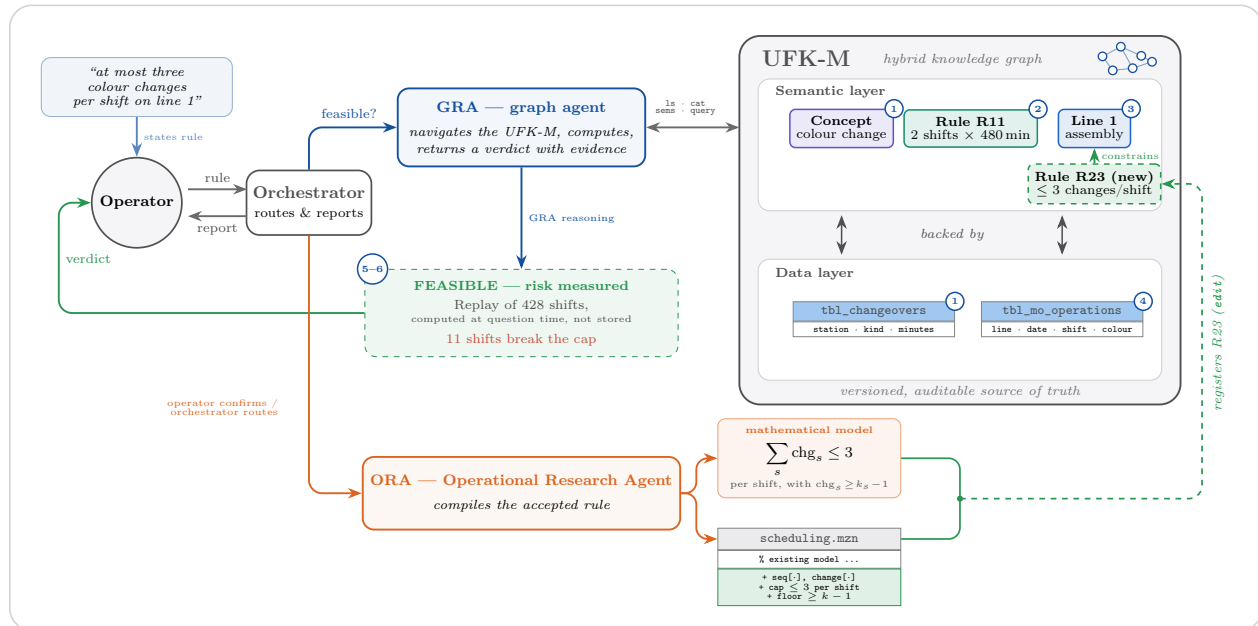

The previous sections measured GRA as a question-answering block. In a
factory, the same block sits inside a wider loop, shown in Figure~\ref{fig:deploy}. An
operator states a rule in plain language; an \textbf{orchestrator} receives
it and asks GRA whether the rule is feasible. GRA gathers its evidence by
navigating the graph with the generic tools of Table~\ref{tab:tools} and returns a verdict
with citations. Deployment adds one write primitive, \texttt{edit}, used only to register approved rules; it is outside the benchmarked toolkit. Accepted rules are passed to \textbf{ORA} (Operational
Research Agent), which turns them into mathematical optimization models and
solver code, and registers them back into the graph as new rule nodes. The
two examples below follow the two possible outcomes: a rule that must be
refused (Figure~\ref{fig:example-refuse}) and a rule that is accepted and compiled (Figure~\ref{fig:example-compile}).

\subsection{Example 1 --- refusing an impossible rule, with evidence}

\textbf{Operator query.} \emph{``Aluminium frames go to welding station~1
or~2 on Monday; station~3 is under maintenance.''} The sentence is clear,
well-formed, and would compile into a valid constraint. It is also
impossible, for two reasons that live in two different places.

\textbf{GRA reasoning.} Asked by the orchestrator whether the rule is
feasible, the agent proceeds in four steps.
\begin{enumerate}[leftmargin=1.5em, itemsep=2pt, topsep=2pt]
  \item \emph{Find what the words point to.} The three welding stations are
        located in the graph, together with the table that records welding
        operations.
  \item \emph{A hidden rule excludes station~1.} Listing the edges of
        station~1 surfaces an older quality rule (node R7 in the graph):
        this station welds carbon frames only, because aluminium dust
        damages carbon parts. The rule speaks about carbon; the request
        speaks about aluminium. The two share no word, but in the graph it sits
        one edge from the station. Station~1 is out; only station~2 remains.
  \item \emph{At standard times, the day barely fits.} Can one station
        absorb the whole Monday? No stored value answers this, so the agent
        computes it: the aluminium frames due that day, multiplied by the
        standard welding time, need 936 minutes, and the working calendar
        (rule R11: two shifts of eight hours) offers 960. On paper, the
        rule passes.
  \item \emph{Measured history says otherwise.} The agent then checks the
        standard against the durations actually recorded over past months:
        real welds run about a third longer than the standard, and each
        switch of material on the station costs extra minutes. Recomputed
        with measured values, the day needs about 1{,}300 minutes, a
        shortfall of more than a full shift.
\end{enumerate}

\textbf{GRA advice.} The rule is refused for two independent reasons: a
conflict with an active quality rule, and a capacity shortfall that only
appears when measured history replaces standard times. The refusal is
returned with its evidence and two repair options: suspend the carbon-only
rule for one day (requires quality approval), or move three orders to
Tuesday (two deliveries become late). Neither the conflict nor the
shortfall was stored anywhere; both were found, or computed, at question
time.

\textbf{Tool-call reasoning trace.} Nine calls, all from Table~\ref{tab:tools}.

\begin{center}
\small
\renewcommand{\arraystretch}{1.25}
\begin{tabularx}{\linewidth}{l X}
\toprule
\textbf{Call} & \textbf{What it does} \\
\midrule
\texttt{sems("welding station")} & locate the three stations in the graph \\
\texttt{ls(welding station 1)}           & list its edges --- the carbon-only rule appears \\
\texttt{cat(rule R7)}            & read it: station~1 welds carbon frames only \\
\texttt{cat(rule R11)}           & daily workload: two shifts, 960 minutes \\
\texttt{query} (SQL)             & Monday load at standard times: 936 minutes \\
\texttt{query} (SQL)             & measured weld durations: a third above standard \\
\texttt{query} (SQL)             & changeover minutes on station~2 \\
\texttt{think}                   & recompute the load: $\approx$1{,}300 for 960 available \\
\texttt{answer}                  & refusal, two repairs, evidence cited \\
\bottomrule
\end{tabularx}
\end{center}

\vspace{0.7cm}

\subsection{Example 2 --- compiling an accepted rule into the scheduling model}
 
\textbf{Operator query.} \emph{``At most three colour changes per shift on
line~1.''} This rule is admissible --- but the scheduling model (MiniZinc code file \textit{scheduling.mzn}) has no
object named ``colour change'', so accepting it takes more than approval:
someone must decide what it means, check it against history, and write it
into the model.

\textbf{GRA reasoning.} Asked whether the rule is feasible, the agent
proceeds in four steps.
\begin{enumerate}[leftmargin=1.5em, itemsep=2pt, topsep=2pt]
  \item \emph{Find what the words point to.} ``Colour change'' refers,
        through a concept node, to the table that logs every changeover;
        ``shift'' refers to the working calendar (rule R11): two shifts
        of eight hours per day.
  \item \emph{Check for conflicts.} Listing what already governs line~1
        surfaces no rule that contradicts the request.
  \item \emph{Replay history.} No stored value can judge this rule. The agent
        therefore needs to count the colours due on each of the 428 shifts in the
        recorded history.
  \item \emph{Quantify the risk.} On 11 of those shifts more than three changes were unavoidable.
        On those days, the rule would have made planning impossible.
\end{enumerate}

\textbf{GRA advice.} Feasible, with a measured seasonal risk. The agent
returns the verdict with two ways to adopt the rule: as a hard limit with
a manual override for peak days, or as a soft constraint penalised in the
objective, which the solver may violate at a bounded cost when a shift
would otherwise be infeasible. The operator confirms the hard form. As in
path~1, the verdict existed nowhere as stored text; it was computed from
the tables at question time.

\textbf{ORA compilation.} Once the operator has approved the rule, ORA
turns it into something the solver can run. It first writes the rule as a
mathematical constraint, then implements it in the solver code: a new
variable for the order of jobs on line~1, a marker for each colour change,
and one line that caps the changes per shift at three. The same rule now
exists in two matching forms --- the mathematical model and the
\texttt{scheduling.mzn} code.

\textbf{Human approval and write-back.} The mathematical formulation and
the exact lines of code are sent to Oplit's OR experts as
a pull request, which they review and approve before anything is merged.
Only then does the accepted rule become node R23 in the graph, linked to
the exact lines of \texttt{scheduling.mzn} and to the validation evidence.

\textbf{Tool-call reasoning trace.} Six calls to reach the verdict. After
the operator confirms and Oplit's operations-research experts approve the
pull request, one edit step 
registers the rule into the UFK-M and closes the loop.

\begin{center}
\small
\renewcommand{\arraystretch}{1.25}
\begin{tabularx}{\linewidth}{l X}
\toprule
\textbf{Call} & \textbf{What it does} \\
\midrule
\texttt{sems("colour change")} & resolve the phrase: a concept node, backed by the changeover table \\
\texttt{cat(rule R11)}         & what a shift is: two shifts of eight hours per day \\
\texttt{ls(line 1)}            & list what already governs the line --- no conflict \\
\texttt{query} (SQL)           & colours due per shift, over the 428 recorded shifts \\
\texttt{think}                 & weigh the result: is the risk acceptable? \\
\texttt{answer}                & feasible, seasonal risk quantified --- the operator confirms \\
\midrule
\emph{ORA}     & formalises the mathematical model, then writes the solver code \\
  \texttt{edit}  & registers rule R23, linked to the new code and the evidence \\
\bottomrule
\end{tabularx}
\end{center}

\section{Conclusion}

A coding agent understands an unfamiliar codebase by navigating it with a
few generic commands rather than reading it whole. This paper showed that a
hybrid industrial knowledge graph admits the same interface: an agent with
seven generic primitives answers analytical questions over a graph it has
never seen, discovering the schema, vocabulary, tables, and join paths
through tool use alone. On models that call tools reliably, this
schema-agnostic GRA outperforms a full-context agent by up to $5.1$\,pp
while reading a quarter to a third of its input. Seeing less, the agent answers better: selective navigation over a structured
substrate beats exhaustive context.

Beyond question answering, GRA supports a harder task. Inside the loop of Figure~\ref{fig:deploy}, it judges whether a new operational rule is feasible, crossing rules, tables, and computed results drawn from both the semantic and data layers. The graph is what makes this tractable: it exposes the paths along which a rule interacts with those already in force, sometimes several hops away, surfacing conflicts and limits that no single system records. A scattered manual investigation becomes a grounded, auditable verdict,
produced with the same tools the agent was benchmarked on.

The next step is ORA. Turning an approved rule into a running constraint
requires translating natural language into a correct mathematical
formulation, then into formal solver code that composes with the
constraints already in force. Future work targets that translation,
making it reliable, verifiable, and grounded in the graph.

\newpage
\bibliographystyle{unsrtnat}
\bibliography{references}

\end{document}